\pdfoutput=1

\documentclass[11pt]{article}

\usepackage[]{acl}

\usepackage{times}
\usepackage{latexsym}

\usepackage[T1]{fontenc}

\usepackage[utf8]{inputenc}

\usepackage{microtype}

\usepackage{latexsym}

\usepackage{colortbl}
\usepackage{booktabs}
\usepackage{multirow}
\usepackage{graphicx}
\usepackage{subcaption}
\usepackage{pgfplots}
\usepackage[ruled,vlined]{algorithm2e}
\usepackage[T1]{fontenc}
\usepackage{diagbox}
\usepackage{amsmath}
\usepackage{todonotes}
\usepackage{lipsum}

\definecolor{light-gray}{gray}{0.9}

\newcommand\blfootnote[1]{%
  \begingroup
  \renewcommand\thefootnote{}\footnote{#1}%
  \addtocounter{footnote}{-1}%
  \endgroup
}

\definecolor{results_red}{RGB}{255,0,0}
\definecolor{results_blue}{RGB}{0,0,255}

\definecolor{table_decrease}{RGB}{255,200,200}
\definecolor{table_increase}{RGB}{200,200,255}

\definecolor{tablebest}{RGB}{192,192,192}

\title{Emotion analysis and detection during COVID-19}

\author{Tiberiu Sosea$^{\clubsuit *} \mbox{  }$
Chau Pham$^{\diamondsuit *} \mbox{  }$
Alexander Tekle$^{\heartsuit}  \mbox{  }$
Cornelia Caragea$^{\clubsuit \dagger}  \mbox{  }$
Junyi Jessy Li$^{\spadesuit \dagger}$ \\ 

  $^{\clubsuit}$Computer Science, University of Illinois at Chicago \\
  $^{\diamondsuit}$Computer Science, Colgate University \\
  $^{\heartsuit}$Computer Science, The University of Texas at Austin \\
  $^{\spadesuit}$Linguistics, The University of Texas at Austin \\

  \texttt{\{tsosea,cornelia\}@uic.edu},
  \texttt{chau@colgate.edu}, \\
  \texttt{alexander.tekle@utexas.edu}, 
  \texttt{jessy@austin.utexas.edu}
}

\begin{document}

\maketitle

\begin{abstract}
Crises such as natural disasters, global pandemics, and social unrest continuously threaten our world and emotionally affect millions of people worldwide in distinct ways. Understanding emotions that people express during large-scale crises helps inform policy makers and first responders about the emotional states of the population as well as provide emotional support to those who need such support. We present {\sc CovidEmo}, $\sim$3K English tweets labeled with emotions and temporally distributed across 18 months. Our analyses reveal the emotional toll caused by COVID-19, and changes of the social narrative and associated emotions over time. Motivated by the time-sensitive nature of crises and the cost of large-scale annotation efforts, we examine how well large pre-trained language models generalize across domains and timeline in the task of perceived emotion prediction in the context of COVID-19. Our analyses suggest that cross-domain information transfers occur, yet there are still significant gaps. We propose semi-supervised learning as a way to bridge this gap, obtaining significantly better performance using unlabeled data from the target domain.\blfootnote{*, $\dagger$: Equal contribution.}
\end{abstract}

\section{Introduction}
Large-scale crises, such as pandemics, natural disasters, and social crises, drastically shift and reshape the physical and mental well-being of millions. Understanding emotions that people increasingly express on social media during large-scale crises can have wide-ranging implications, from promoting a deeper understanding of the society to informing policy makers and first responders about the emotional states of the population~\cite{dennis2006making,fraustino2012social}. In Natural Language Processing (NLP), multiple datasets have been proposed to detect emotions on social media~\cite{mohammad-2012-emotional,wang2012harnessing,mohammad2015using,volkova-bachrach-2016-inferring,abdul-mageed-ungar-2017-emonet,demszky2020goemotions}, including from hurricane disasters~\cite{Schulz2013AFS,desai2020detecting}.

We explore the detection of perceived fine-grained emotion during the COVID-19 pandemic to answer two research questions. First, from a \textbf{social} point of view,  each crisis is situated in its own unique social context \cite{palen2016crisis}, triggering distinct emotions, and impacting different populations in vastly distinct ways. COVID-19 is a crisis that has dominated the world stage and influenced every aspect of human life. What are the emotions expressed through social media, and how do they change over time? Second, from a \textbf{system} point of view, modern data-driven emotion prediction systems are trained on large, annotated datasets. How well can models learn from existing resources since timely annotation of fine-grained emotions can be costly to accumulate as new crises arise, and how well do models generalize as a crisis unfolds through different stages?

\begin{table}
\centering
\small
\begin{tabular}{p{0.75\linewidth} p{0.15\linewidth}}
 \toprule
  \rowcolor{light-gray}
 \texttt{<USER>} Please resign, you are the master of misleading who started politicizing the public health crisis. You are a part of the problems the world is facing! & anger, disgust, sadness \\ \midrule
 'Perfect storm': Haiti COVID-19 peak set to collide with hurricanes. \texttt{<URL>} & fear, sadness \\ \midrule
  \rowcolor{light-gray}
 The German government is taking all kind of measures to protect its people while the Dutch government does not care about their people \#corona & surprise, trust, anger\\
 \bottomrule
\end{tabular}
\caption{\label{tb:sample} Examples from {\sc CovidEmo} annotated with the Plutchik-8 emotions.}
\label{tab:examples}
\end{table}

To answer these questions, we introduce {\sc CovidEmo}, a dataset of \textasciitilde3K tweets in English annotated with Plutchik-8 emotions~\cite{plutchik2001nature}; examples shown in Table~\ref{tab:examples}. This dataset provides an ideal test bed to examine how well modern NLP models generalize across domains and crises in the task of perceived emotion prediction. Moreover, {\sc CovidEmo} is temporally distributed across $18$ months, which 
enables the exploration of distributional shifts that occurred from the start of the pandemic. Our analysis reveals that the co-occurrence and distribution of emotions are drastically different from natural disasters such as hurricanes~\cite{desai2020detecting}. However, while \citet{desai2020detecting} pointed out that emotion distributions are fairly consistent across hurricanes, in {\sc CovidEmo} we observe a different phenomenon: as COVID-19 progressed, we note considerable distributional shifts both in the lexical and the emotion label space. Additionally, we found that politically related words are more likely to associate with negative emotion, while vaccine-related words are more likely to associate with positive ones.

We carry out a comprehensive set of experiments that evaluate model generalizability under domain shift. 
Experimenting with large-scale pre-trained language models including BERT~\cite{devlin2019bert}, TweetBERT~\cite{bertweet}, and COVID-Twitter-BERT~\cite{muller2020covid},
we find that directly applying models trained on other emotion datasets to {\sc CovidEmo} leads to poor overall performance, indicating considerable domain gaps. Our analysis also reveals two surprising findings: \textbf{1)} Performing direct transfer from a general emotion dataset such as GoEmotions \cite{demszky2020goemotions} attains better performance compared to transferring information from a disaster-specialized corpus such as HurricaneEMO \cite{desai2020detecting}, indicating the vast differences across crises. \textbf{2)} Besides the inter-domain gaps observed, we note in-domain model performance gaps along the temporal dimension as well. Specifically, we find that training a model on the first 6 months of our data and testing on the last 6 months obtains a 2\% decrease in F-1 score compared to using training and testing data from the same timeframe (last six months).

Finally, we investigate methods to bridge both the inter-domain and the in-domain temporal gaps. We motivate the importance of lowering these gaps: first, due to the time-critical, dynamic nature of disasters such as COVID-19, the time needed to acquire labeled data might severely impact the early-risk assessment capabilities of the authorities and slow the relief response. Second, labeling data for every potential disaster is not feasible in terms of annotation costs. To this end, we leverage Noisy Student Training \cite{xie2020self}, a semi-supervised learning technique utilizing the readily available COVID-19 unlabeled data, and the non-COVID labeled data, to obtain a better emotion detection model. 
This improves the performance of the vanilla models significantly, by 1.5\% on average.

We summarize our contributions as follows: \textbf{1)} We introduce {\sc CovidEmo}, an emotion corpus containing $\sim$3K tweets streamed during the COVID-19 pandemic, which enables the exploration of  model generalization across  domains, as well as between different time periods of the same domain. \textbf{2)} We perform a comprehensive analysis of emotion expression in {\sc CovidEmo}, indicating various particularities and comparing our corpus with other datasets in the literature. \textbf{3)} We observe considerable domain gaps and offer potential explanations into why models struggle to transfer information. \textbf{4)} We bridge these gaps using semi-supervised learning. 
We will release our data and models upon publication.

\begin{table*}
\centering
\small
\begin{tabular}{p{0.08\linewidth} | p{0.85\linewidth}}
 \toprule
 \textbf{Emotion} & \textbf{Content words/Hashtags} \\
 \midrule
 \rowcolor{light-gray}
 disgust & \textbf{Content words: } disgusting, fucking, million, trump, dead, shit, president, america, china, done \newline \textbf{Hashtags: }\#hongkong, \#gop, \#factsmatter, \#ccp, \#china, \#wuhan, \#covid19 \\ 
 anger & \textbf{Content words: } fuck, evil, bullshit, stupid, idiot, damn, obama, church, lying \newline \textbf{Hashtags: } \#marr, \#covidiots, \#trumpvirus, \#torycorruption, \#skynews, \#qanon, \#nh, \#jacksonville, \#gop, \#factsmatter \\ 
 \rowcolor{light-gray}
 fear & \textbf{Content words: } scared, exam, dangerous, infected, confirmed, worse, sir, wuhan, risk, rate \newline \textbf{Hashtags: } \#stopcovidlies, \#jeeneet, \#antistudentmodigovt, \#health, \#wuhan, \#china, \#stayhome, \#covid19 \\ 
 sadness & \textbf{Content words: } sad, cry, died, suffering, toll, record, sorry, feel, tested, facing \newline \textbf{Hashtags: } \#notmychild, \#quarantine, \#rip, \#pregnant, \#italy, \#healthcare, \#freepalestine, \#askktr, \#wuhan, \#vaccine \\ 
 \rowcolor{light-gray}
 anticipation & \textbf{Content words: } effort, christmas, available, join, start, future, vaccination, vaccinated, coming, open \newline \textbf{Hashtags: } \#stayhomestaysafe, \#pregnant, \#postponeinicet, \#nyc, \#launchzone, \#fred2020, \#cow, \#whatshappeninginmyanmar, \#ethereum, \#bcpoli\\ 
 trust & \textbf{Content words: } working, support, safe, help, say, being, world, vaccine, good, more \newline \textbf{Hashtags: } \#stayhome, \#staysafe, \#covid19, \#lockdown, \#china \\ 
 \rowcolor{light-gray}
 joy & \textbf{Content words: } grateful, beautiful, thanks, happy, love, great, little, morning, good \newline \textbf{Hashtags: } \#taiwan, \#innovation, \#breaking, \#staysafe, \#stayathome, \#stayhome, \#wearamask, \#lockdown, \#covid19\\ 
 surprise & \textbf{Content words: } believe, year,  lockdown, new, china, virus, day, america, covid19, get \newline \textbf{Hashtags: } \#china, \#covid19\\ 
 \bottomrule
\end{tabular}
\caption{\label{tb:lexical-analysis} Content words and hashtags most associated with each Plutchik-8 emotion.}
\end{table*}

\section{Data}

\subsection{Data collection}

\paragraph{Preprocessing.}
We sample $129,820$ English tweets from
\citet{emilychendata}'s  ongoing collection of tweets related to the COVID-19 pandemic, starting from January 2020 until June 2021\footnote{We use the Twarc software to obtain the tweet texts, and FastText~\cite{fasttext} for language identification.}.
Our sampling strategy involves selecting an equal number of tweets each month in the time period mentioned above. The tweets are anonymized by replacing twitter usernames with \texttt{<USER>} and links with \texttt{<URL>}, following \citet{cachola-etal-2018-expressively}. Additionally, prior work found that even in disaster contexts, the fraction of tweets expressing an emotion is small~\cite{desai2020detecting}, thus annotating randomly sampled tweets would be costly and unproductive. Therefore, we follow their work 
to obtain tweets that are more likely to contain emotions for annotation. Concretely, we ensure 
that each tweet encompasses at least one word from EmoLex \cite{emolex}, a lexicon of \textasciitilde10K words in various languages annotated with emotion labels. After this filtering process, we obtain 89,274 tweets. As stated in \citet{desai2020detecting}, this filtering is soft, i.e., does not filter out tweets with weak or implicit emotions.

\paragraph{Annotation and quality control.}
We randomly sample $5,500$ tweets from this data and use  Amazon Mechanical Turk to crowdsource Plutchik-8 emotions: \emph{anger, anticipation, joy, trust, fear, surprise, sadness, disgust}.
We allow multiple selection, as well as a \emph{none of the above} option in case no emotion is perceived. During the annotation process, we determine the inter-annotator agreement using the Plutchik Emotion Agreement (PEA) metric that take into account emotion proximity on the Plutchik wheel \cite{desai2020detecting}.

We use a qualification process for quality control and training. Specifically, two members of our research team annotated a small set of tweets, from which we selected 20 examples where both annotators agree on the emotions. We qualify workers whose annotations attain high agreement with ours (PEA$>$75.00) calculated against our annotations. This results in a highly capable pool of workers for the main task. 
Additionally,
we exclude annotations from workers who have very poor agreement with others~\cite{cachola-etal-2018-expressively,desai2020detecting} (those whose PEA scores are below the 80th percentile compared to others). 
Each tweet has at least 2 annotations after filtering.

We aggregate labels such that an emotion is considered present if \emph{at least two workers} perceived the emotion. This resulted in $2,847$ tweets in {\sc CovidEmo} with an average, per-worker PEA score of $84.05$, indicating high inter-annotator agreement.

\begin{figure}[t]
\centering
\includegraphics[width=\linewidth]{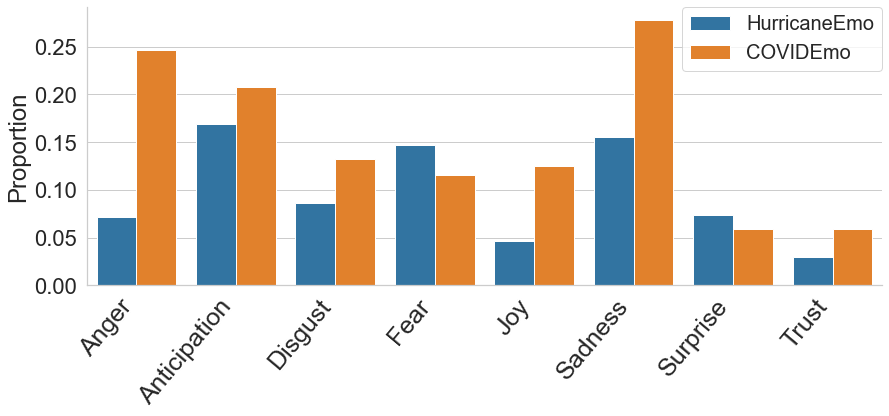}
\caption{Emotion distribution of two types of crises: hurricanes and the COVID-19 pandemic.}
\label{fig1}
\end{figure}

\subsection{Analysis}\label{sec:data_analysis}

\paragraph{Emotion distribution.}
We show the general distribution of Plutchik-8 emotions in {\sc CovidEmo} in Figure~\ref{fig1}. 
We note that the percentage of negative emotions (\emph{disgust, anger, fear, sadness}) is much higher than that of positive emotions (\emph{trust, joy}), consistent of the emotional toll of COVID-19. Next, we draw comparisons between the emotion distribution in COVID-19 and that of natural disasters, specifically HurricaneEmo \cite{desai2020detecting}, shown in Figure~\ref{fig1}. We make a few observations: First, the tweets in \textsc{CovidEmo} contain a higher emotion proportion across six out of the eight total emotions, indicating that COVID-19 prompted an increased multi-label emotional response compared to natural disasters.
Second, 
the sadness emotion is almost twice more represented in \textsc{CovidEmo} compared to HurricaneEmo, whereas we see as much as a four-fold increase in the representation of anger. Finally, we observe that \emph{anticipation} is much more prevalent in HurricaneEmo and a lot less frequent in the pandemic, which matched the COVID-19 reality that it is hard to anticipate events/facts.

\begin{figure}[t]
\centering
\includegraphics[width=\linewidth]{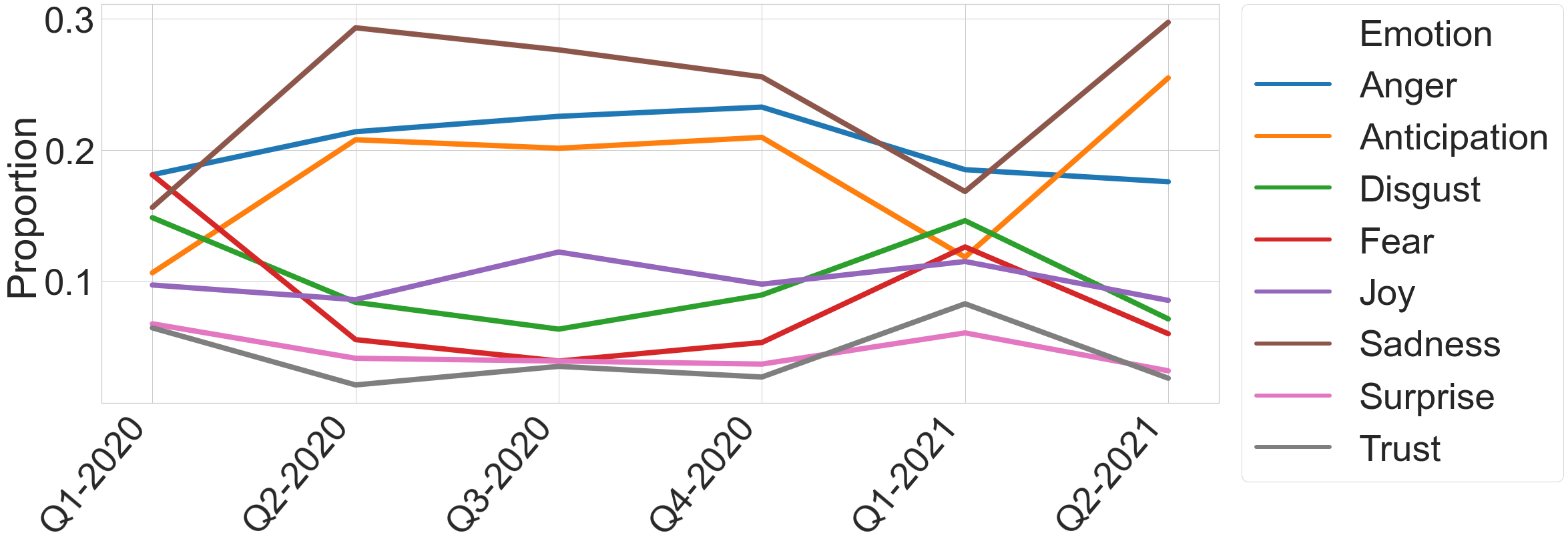}
\caption{Emotion distribution in {\sc CovidEmo} over time (by quarter).}
\label{fig2}
\end{figure}

We also show \textbf{emotion distribution across time} in Figure \ref{fig2}, obtained grouping the tweets by quarter (e.g., Q1-2020 encompasses the first three months of 2020). We observe that the label distribution varies significantly from quarter to quarter, denoting potential changes in the discussion topics or the overall feelings of the masses. Notably, we note proportion variations as high as 12\% in consecutive quarters. For instance, the proportion of the sadness emotion increases by as much as 12\% in the second quarter of 2020 compared to the first quarter. Moreover, we see the opposite trend in the fear emotion, whose proportion decreases by 10\% percent. One potential explanation could be that the first shock that COVID-19 produced enacted fear into people (Q1 2020). However, as people started to get accustomed to the lockdown, the fear slowly turned into sadness. 

\begin{figure}[t]
\centering
\includegraphics[width=0.7\linewidth]{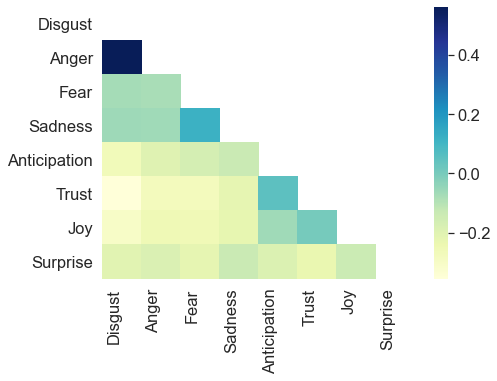}
\caption{Emotion co-occurrence in {\sc CovidEmo}.}
\label{fig3}
\end{figure}

\paragraph{Emotion co-occurrence.}
Figure~\ref{fig3} depicts how emotions co-occur with one another in {\sc CovidEmo}. For each emotion pair, we compute the Pearson correlation coefficient. Overall, we observe stronger correlation between emotions in the same positive/negative categories. For example, (\emph{anger, disgust}) and (\emph{sadness, fear}) appear much more frequently than (\emph{anger, anticipation}) and (\emph{anger, joy}).
Table \ref{tb:sample} shows samples from {\sc CovidEmo} with multiple emotions perceived. 
Notably, in many cases lexical cues alone cannot account for the emotions, as evident in the second example. Although the word ``perfect'' suggest optimism, the annotations are nowhere near positive. In the third tweet, there is a co-occurrence of polarizing emotion because the tweet deals with a positive and a negative situation at the same time. 

\begin{table*}[t!]
\centering
\setlength{\tabcolsep}{18pt}
\small
\centering
\begin{tabular}{cccccc}
\toprule
  \textsc{Q1 2020} & \textsc{Q2 2020} & \textsc{Q3 2020} & \textsc{Q4 2020} & \textsc{Q1 2021} & \textsc{Q2 2021} \\
  \midrule
  china  & pandemic & test & vaccine & biden & india\\
  virus & lockdown & school & health & vaccine & vaccination\\
  outbreak & stimulus & government & trump & lockdown & help\\
  trump & world & trump & biden & death &  death\\
  health & health & death & american & school & country\\
  spread & test & student & positive & stimulus & government\\
\bottomrule
\end{tabular}
\caption{Words associated with the most prominent topics by quarter identified using LDA.}
\label{tab:lda-result}
\end{table*}
\paragraph{Lexical analysis.}
To understand better what topics or events are associated with each emotion, we perform a lexical analysis to examine the co-occurrence between content words (nouns, verbs, adjectives and adverbs), hashtags and perceived emotions. 
In particular, we calculate the log odds ratios ($\log(P(w|e)/P(w|\neg e)$)~\cite{nye2015identification} with a frequency threshold of 10 for lemmatized content words and 2 for hashtags. 
Table \ref{tb:lexical-analysis} shows the highest ranked content words and hashtags for each emotion category. We notice that politically or country-oriented words are more likely to associate with negative emotions \textit{(president, america, china)}, while vaccine-related words are more likely to associate with positive emotions.

\paragraph{Topical variations.} We aim to understand the temporal gaps in \textsc{CovidEmo} by examining what topics have been discusses during the pandemic. In this study, we use all the collected unlabeled examples and group them by quarters. For each quarter, we remove the punctuation and stopwords, convert the words to lowercase and perform lemmatization. Next, we use latent Dirichlet allocation (LDA) \cite{10.5555/944919.944937} to extract the most prominent topics in our corpus. We show the unigrams corresponding to the most discussed topics in Table \ref{tab:lda-result}. Interestingly, we can clearly trace major events and examine the topical drifts that occurred during COVID-19. For instance, in the first quarter of 2020, we see words such as \emph{china}, \emph{outbreak} or \emph{spread}, which denote the main events happening at the time \cite{q1-who}. Next, in Q2 2020 we trace events such as the arrival of the first stimulus check \cite{q2-stimulus}, testing, and the aftermath of COVID-19 being declared a pandemic  \cite{q2-pandemic}. Throughout Q3, we notice debate over school reopening as the Fall semester was approaching \cite{q3-school}. In Q4, we can discern topics such as the US election \cite{q4-election} and first discussions about the vaccine \cite{q4-vaccine}. Next, in Q1 2021 we observe discussions about the next stimulus check \cite{q1-stimulus}, while in Q2 2021 conversations about the record-breaking number of COVID-19 cases in India arise \cite{q2-india}. All these variations of the topics discussed during the pandemic could explain the performance gaps in our temporal adaptation experiments. 

\begin{table}[t!]
\centering
\setlength{\tabcolsep}{4.5pt}
\small
\centering
\begin{tabular}{rccccccccc}
\toprule
 & \textsc{ang} & \textsc{ant} & \textsc{dis} & \textsc{fea} & \textsc{joy} & \textsc{sad} & \textsc{sur} & \textsc{tru}  \\ \midrule
\textsc{dev} & $327$ & $296$ & $163$ & $179$ & $186$ & $388$ & $86$ & $89$ \\
\textsc{test} & $374$ & $296$ & $214$ & $149$ & $170$ & $403$ & $83$ & $78$ \\
\bottomrule
\end{tabular}
\caption{Validation and test set splits for eight Plutchik-8 emotions, including including anger (ang), anticipation (ant), disapproval (dis), fear (fea), joy, sadness (sad), surprise (sur), trust (tru).}
\label{tab:test-val-data}
\end{table}

\subsection{Benchmark Dataset} 
To enhance reproducibility and aid the progress on understanding the expression of emotion in the COVID-19 context, we use {\sc CovidEmo} as a benchmark dataset for perceived emotions.
We split our data into a development and testing split, as shown in Table \ref{tab:test-val-data}. We also note that the data is evenly distributed across the time axis, with an equal number of $158$ tweets  for each of the $18$ months that our dataset spans. As mentioned previously, disasters are time-critical events, and since our goal is to examine the emergence of such disasters, we mainly focus on domain adaptation techniques, hence we omit creating a training set.

\section{Domain Transfer Assessment}

Using {\sc CovidEmo}, we evaluate the ability of modern NLP models to transfer information from existing sources with annotated emotions in an inter-domain setting for perceived emotion detection, and if models generalize temporally in the same larger context (in-domain temporal transfer). 

\subsection{Our Framework} 
We consider a dataset $\mathcal{S}$ labeled with emotions, and another collection of labeled examples $\mathcal{T}$ from a different domain. We aim to assess how well large pre-trained language models can transfer information from the domain of $\mathcal{S}$ to the domain of $\mathcal{T}$. To this end, we train our models on $\mathcal{S}$, then evaluate the performance on the test set of $\mathcal{T}$. In our framework, $\mathcal{T}$ is {\sc CovidEmo} for the inter-domain experiments, or a temporal slice of {\sc CovidEmo} for the temporal experiments. Due to the uneven label distribution and the multi-label nature of the data, we develop binary classifiers for each emotion following \citet{desai2020detecting}.

\begin{table*}[t!]
\centering
\setlength{\tabcolsep}{7.6pt}
\small
\centering
\begin{tabular}{r|cccccccc|c}
\toprule
\textsc{model} &  \textsc{ang} & \textsc{ant} & \textsc{dis} & \textsc{fea} & \textsc{joy} & \textsc{sad} & \textsc{sur} & \textsc{tru} & \textsc{avg} \\
\midrule
\textsc{bert-goemotions} & $0.735$ & $0.589$ & $0.624$ & $0.625$ & $0.722$ & $0.687$ & $0.588$ & $0.540$ & $0.635$ \\ 
\textsc{bert-hurricaneemo} & $0.592$ & $0.339$ & $0.563$ & $0.398$ & $0.385$ & $0.467$ & $0.403$ & $0.347$ & $0.433$ \\
\midrule
\textsc{tweetbert-goemotions} & $0.752$ & $0.534$ & $0.631$ & $0.629$ & $0.709$ & $0.708$ & $0.624$ & $0.537$ & $0.637$ \\ 
\textsc{tweetbert-hurricaneemo} & $0.677$ & $0.346$ & $0.540$ & $0.311$ & $0.299$ & $0.494$ & $0.354$ & $0.418$ & $0.435$ \\ 
\midrule
\textsc{ctbert-goemotions} & $0.735$ & $0.577$ & $0.629$ & $0.644$ & $0.725$ & $0.717$ & $0.617$ & $0.520$ & $0.644^{\dagger}$ \\ 
\textsc{ctbert-hurricaneemo} & $0.655$ & $0.366$ & $0.471$ & $0.311$ & $0.341$ & $0.447$ & $0.243$ & $0.349$ & $0.406$ \\ 
\midrule
\textsc{emolex} & $0.57$ & $0.517$ & $0.547$ & $0.551$ & $0.543$ & $0.560$ & $0.458$ & $0.414$ & $0.504$ \\ 
\bottomrule
\end{tabular}
\caption{Direct transfer Macro F-1 scores using BERT \cite{devlin2019bert} base uncased model (\textsc{bert}-*), TweetBERT \cite{bertweet} (\textsc{tweetbert}-*) and Covid-Twitter-BERT (\textsc{ctbert}-*). The results in this table are average F-1s across $5$ different runs. We assert significance$^{\dagger}$ if $p < 0.05$ under a paired-t test with the vanilla BERT model.}
\label{tab:direct_transfer}
\end{table*}

\begin{table*}[t!]
\centering
\setlength{\tabcolsep}{11pt}
\small
\centering
\begin{tabular}{r|cccccccc|c}
\toprule
\textsc{model} &  \textsc{ang} & \textsc{ant} & \textsc{dis} & \textsc{fea} & \textsc{joy} & \textsc{sad} & \textsc{sur} & \textsc{tru} & \textsc{avg} \\
\midrule
\textsc{ctbert-}$\mathcal{F}_{tr}$ & $0.762$ & $0.485$ & $0.534$ & $0.661$ & $0.705$ & $0.673$ & $0.492$ & $0.492$ & $0.600$ \\
\textsc{ctbert-}$\mathcal{L}_{tr}$ & $0.769$ & $0.631$ & $0.498$ & $0.668$ &  $0.781$ & $0.724$ & $0.493$ & $0.502$ & $0.633$$^{\dagger}$\\ 
\midrule
\end{tabular}
\caption{Macro F-1 scores using in-domain temporal adaptation. The \textsc{ctbert-}$\mathcal{L}_{tr}$ improvements are statistically significant$^{\dagger}$.}
\label{tab:temporal_gap}
\end{table*}

\paragraph{Methods.} Motivated by the tremendous success of large pre-trained masked language models, 
we use the following models: \textbf{1)} BERT \cite{devlin2019bert} base uncased model trained on Wikipedia and BookCorpus \cite{zhu2015aligning}, \textbf{2)} BertTweet \cite{bertweet} model trained on 850M english tweets, and \textbf{3)} COVID-Twitter-BERT (CT-BERT) \cite{muller2020covid} trained on 97M tweets. Additionally, we also employ a basic lexicon-based classification approach, \textbf{4)} EmoLex \cite{emolex} is the word-associated lexicon mentioned previously in the paper. In this approach, if a tweet contains a word annotated with an emotion \emph{e} in EmoLex, then we assign \emph{e} as a label for the tweet.  

\paragraph{Experimental setup.} We perform all our experiments on an Nvidia P100 GPU. To report the performance, we average the F-1s of 5 different runs and report the average value. We present in Appendix A detailed information about the hyperparameters used for the best models. Additionally, in Appendix B, we indicate the hyperparameter search space explored, as well as model running times.

\subsection{Inter-domain Transfer}\label{sec:inter-domain-transfer}
Our first domain transfer assessment explores how well emotion detection models trained outside our domain generalize to the COVID context.
We consider two well-established datasets for training. First, we experiment with GoEmotions \cite{demszky2020goemotions}, a dataset from the general Reddit domain annotated with 28 emotions and the neutral class. The emotion space in GoEmotions differs slightly from our Plutchik-8 setup, hence we perform a mapping\footnote{GoEmotions Mapping: Anger → Anger, Disgust → Disgust, Joy → Joy, Sadness → Sadness, Fear → Fear, Nervousness, Desire → Anticipation, Surprise → Surprise, Admiration → Trust.} between the emotions in GoEmotions and the Plutchik-8 emotions. Second, we use HurricaneEmo \cite{desai2020detecting}, a Twitter dataset collected from natural disasters such as hurricanes and labeled with fine-grained emotions. HurricaneEmo provides Plutchik-8 labels.

\paragraph{Results.} We show the results obtained in Table \ref{tab:direct_transfer}. Here, we denote by \emph{M}-\emph{DS} the model \emph{M} trained on dataset \emph{DS} and tested on \textsc{CovidEmo}. We emphasize a surprising finding: \textbf{models trained on a general domain (GoEmotions) generalize better on \textsc{CovidEmo} compared to models trained on natural disasters such as hurricanes (HurricaneEmo).} In fact, the performance gaps between GoEmotions and HurricaneEmo are vast, and we see as much as 0.20 differences in average macro F-1. At the same time, we note that our basic lexicon-based Emolex approach outperforms the HurricaneEmo transfer models. This result hints to a sizeable divergence between crises such as hurricanes and COVID-19. The CT-BERT model improves the performance by 1\%  on average (with statistical significance), compared to TweetBERT which only obtained marginal improvements. Although both are trained on Twitter data,
we postulate that CT-BERT 
likely benefited from
COVID-related biases that the model manages to leverage.

\begin{table*}[t!]
\centering
\setlength{\tabcolsep}{6pt}
\small
\centering
\begin{tabular}{r|ccc|ccc}
\toprule
& \multicolumn{3}{c}{Cosine Similarity} & \multicolumn{3}{c}{Jensen-Shannon Divergence} \\
\midrule
 & \textsc{Covid} & \textsc{GoEmotions} & \textsc{HurricaneEMO} & \textsc{Covid} & \textsc{GoEmotions} & \textsc{HurricaneEMO} \\ 
\textsc{Covid} & $1.0$ & $0.346$ & $0.243$ & $0.0$ & &  \\
\textsc{GoEmotions} & & $1.0$ & $0.378$ & $0.312$ & $0.0$ & \\
\textsc{HurricaneEMO} & & & $1.0$ & $0.351$ & $0.374$ & $0.0$ \\
\bottomrule
\end{tabular}
\caption{Cosine similarities and Jensen-Shannon divergence of word distributions between GoEmotions \cite{demszky2020goemotions}, HurricaneEMO \cite{desai2020detecting}, and \textsc{CovidEMO}.}
\label{tab:results_similarities}
\end{table*}

\subsection{In-domain Temporal Transfer}

\textsc{CovidEmo} spans a large period of time (18 months) marked by substantial narrative shifts in the society (Table~\ref{tab:lda-result}). Thus we investigate potential distributional shifts across the temporal dimension. Specifically, we aim to analyze how well models trained on past COVID-19 data generalize to a fresh batch of new data. To this end, we stage the following setup: First, we accumulate the subsets $\mathcal{F}$ and $\mathcal{L}$ corresponding to the initial six months and the last six months respectively.
Denoting the development and test sets of {\sc CovidEmo} as $\mathcal{C}_{tr}$ and $\mathcal{C}_{ts}$, we create additional sets $\mathcal{L}_{tr}=\mathcal{L}\cap\mathcal{C}_{tr}$ and $\mathcal{L}_{ts}=\mathcal{L}\cap\mathcal{C}_{ts}$. Additionally, we randomly subsample $\mathcal{F}_{tr} \subset \mathcal{F}$ such that $|\mathcal{F}_{tr}| = |\mathcal{L}_{tr}|$, where $|.|$ denotes the size of a set. In this setting, we compare training on $\mathcal{F}_{tr}$ and testing on $\mathcal{L}_{ts}$ vs.\ training on $\mathcal{L}_{tr}$ and testing on $\mathcal{L}_{ts}$. In other words, we investigate whether model performance on {\sc CovidEmo} decrease as time passes. Here we experiment with CT-BERT \cite{muller2020covid} (since it achieved better performance in Section~\ref{sec:inter-domain-transfer}).

\paragraph{Results.} Table \ref{tab:temporal_gap} shows that \textbf{the models trained on the same time period as the testing data outperforms the model trained on a different timeframe significantly}, obtaining a Macro F-1 increase of 3.3\% on average. Notably, we observe improvements as high as 7.6\% in F-1 on joy and 14.6\% on anticipation. Intuitively, 
since the model is trained on the same temporal distribution as the test set, and anticipation is closely related to ongoing events (i.e., people usually anticipate certain events), it is extremely probable that the model has been trained on similar events in the training set, so the model easily recognizes the emotion.

\section{Understanding Domain Gaps}

The previous section exposed significant inter-domain and temporal gaps leading to poor transfers of information between these domains. 
In this section, we 
aim to answer the following questions: Why does GoEmotions transfer better than HurricaneEmo, even though the latter is a disaster-centric dataset? How did data distribution shift during the pandemic?
We hope that our insights can spur further research into bridging these gaps. 
In Section~\ref{sec:da}, we propose semi-supervised learning as a method to build better transfer learning models.

\paragraph{Inter-domain gaps.} To answer the first question, we analyze the lexical differences between GoEmotions, HurricaneEmo, and \textsc{CovidEmo}. In order to obtain more accurate comparisons in terms of the larger vocabulary, we use unlabeled data for HurricaneEmo and \textsc{CovidEmo} to 
match the number of examples in GoEmotions (\textasciitilde$60$K).
Table~\ref{tab:results_similarities} shows the cosine similarity and the Jensen-Shannon divergence for the frequency distribution of all content words (lower-cased and lemmatized) across the three datasets.
Interestingly, the {\sc CovidEmo} distribution is significantly closer to GoEmotions compared to HurricaneEmo: the cosine similarity is substantially lower ($0.243$ vs. $0.346$) while the divergence is larger ($0.312$ vs. $0.351$). Moreover, the HurricaneEmo distribution diverges even more from GoEmotions compared to COVID-19. These findings hint that although HurricaneEmo is closer to {\sc CovidEmo} than to a general domain, the COVID-19 context is significantly more correlated with a general domain than a natural disaster one, likely due to the wide impact COVID-19 has had and a more social nature of the crisis. These findings could also explain why there are large gaps in performance between HurricaneEmo and GoEmotions transfers.

\begin{figure}[t]
\small
\centering
\includegraphics[width=\linewidth]{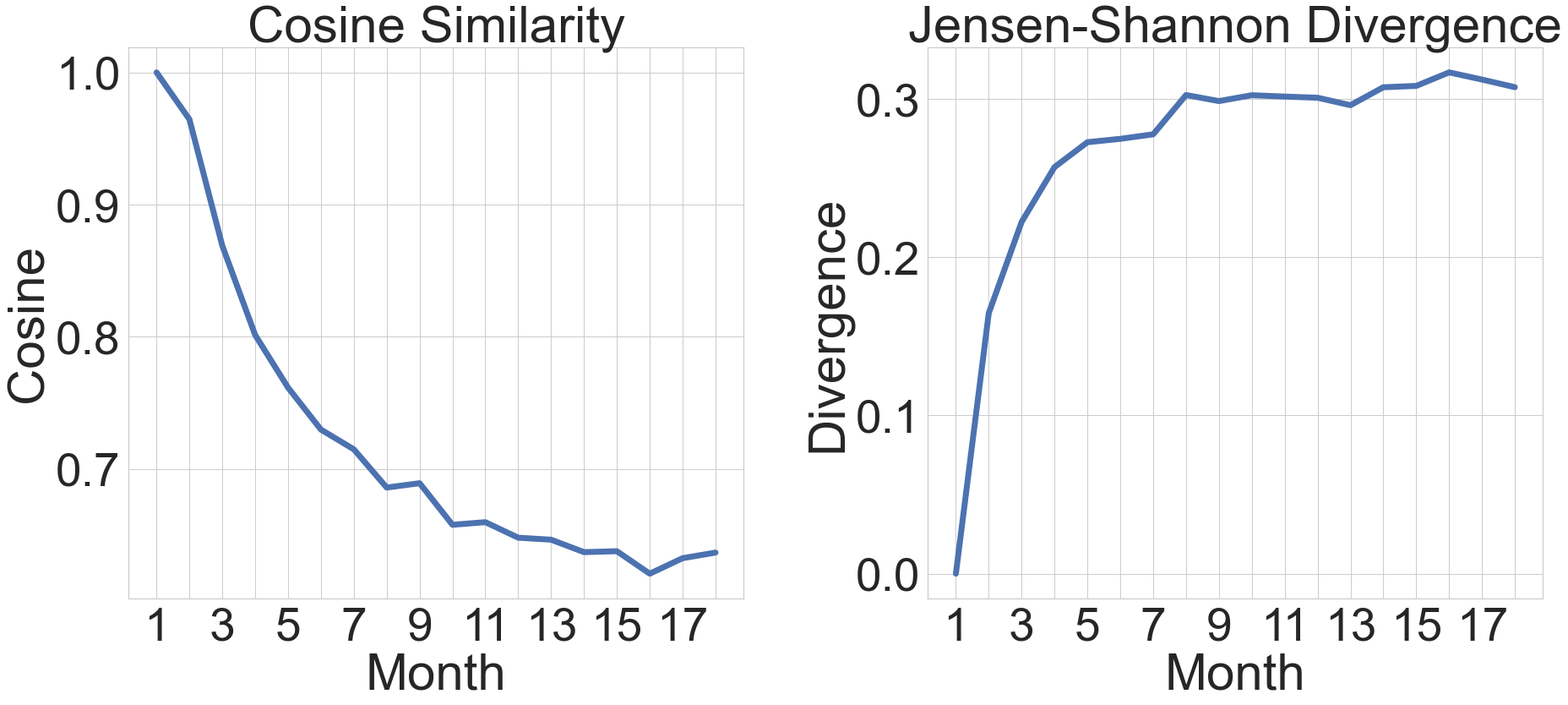}
\caption{Cosine similarities and Jensen-Shannon divergence between the first month of COVID-19 and each subsequent month.}
\label{corpus_similarities}
\end{figure}

\paragraph{In-domain temporal gaps} 
In Section~\ref{sec:data_analysis}, we revealed that the label distribution and topics discussed during COVID-19 has shifted over time.
To consolidate these analyses,
we carry out an additional experiment that captures distributional shifts in vocabulary. 
In Figure~\ref{corpus_similarities} we show the cosine similarities and Jensen-Shannon divergence for the frequency distributions of content words (lower-cased and lemmatized) for unlabeled tweets spanning the 18 months in our data.
As time passes, we observe a constant shift in the lexical distribution of the tweets. Concretely, while the cosine similarity between the first and the second month of COVID-19 is 0.97, by the end of the $18^{th}$ month this value decreases significantly, getting as low as 0.63. We observe the same phenomenon in the divergence of the distributions as well. These findings emphasize the considerable temporal gaps found in long-lasting disasters such as COVID-19, and that temporal slices of the tweets can diverge significantly even though they originate from the same domain.

\begin{table*}[t!]
\centering
\setlength{\tabcolsep}{8pt}
\small
\centering
\begin{tabular}{r|cccccccc|c}
\toprule
\textsc{model} &  \textsc{ang} & \textsc{ant} & \textsc{dis} & \textsc{fea} & \textsc{joy} & \textsc{sad} & \textsc{sur} & \textsc{tru} & \textsc{avg} \\
\midrule
\textsc{ctbert-goemotions} & $0.735$ & $0.577$ & $0.629$ & $0.644$ & $0.725$ & $0.717$ & $0.617$ & $0.520$ & $0.644$ \\ 
\textsc{ctbert-goemotions-ssl} & $0.741$ & $0.554$ & $0.657$ & $0.651$ & $0.741$ & $0.726$ & $0.632$ & $0.532$ & $0.654$$^{\dagger}$ \\ 
\midrule
\textsc{ctbert-}$\mathcal{F}_{tr}$ & $0.762$ & $0.485$ & $0.534$ & $0.661$ & $0.705$ & $0.673$ & $0.492$ & $0.492$ & $0.600$ \\
\textsc{ctbert-}$\mathcal{F}_{tr}$\textsc{-ssl} & $0.771$ & $0.501$ & $0.531$ & $0.711$ & $0.711$ & $0.671$ & $0.538$ & $0.501$ & $0.617$$^{\dagger}$ \\
\textsc{ctbert-}$\mathcal{L}_{tr}$ & $0.769$ & $0.631$ & $0.498$ & $0.668$ &  $0.781$ & $0.724$ & $0.493$ & $0.502$ & $0.633$$^{\dagger}$\\ 
\bottomrule
\end{tabular}
\caption{Macro F-1 scores using inter-domain adaptation (first block), in-domain temporal adaption (second block), and of our best performing models using Noisy Student training \cite{xie2020self}. We assert significance$^{\dagger}$ if $p < 0.05$ under a paired-t test with base model (\textsc{ctbert-goemotions} for inter-domain transfers and \textsc{ctbert-}$\mathcal{F}_{tr}$ for temporal transfers.)}
\label{tab:ssl}
\end{table*}

\section{Bridging the Gaps Between Domains}\label{sec:da} 

As crises such as COVID-19 strike, large amounts of user-generated content are produced on social sites. However, due to the nature of disasters unfolding rapidly and the high costs needed for annotation, immediately obtaining labeled data from the ongoing disaster might prove infeasible. However, rapid understanding of such events is critical for rapid risk assessment and effective resource allocations. Therefore, we cannot rely on obtaining large quantities of labeled data, and we require effective domain adaptation techniques which can leverage labeled data from outside the disaster domain. However, we emphasized previously that models typically have a hard time effectively transferring information for emotion detection. We argue that even though we cannot timely obtain labels for the ongoing disaster, we can still use the large amounts of unlabeled user-generated Tweets to build better domain adaptation models. To this end, we experiment with semi-supervised learning.

\paragraph{Method.} Noisy student training \cite{xie2020self} is an approach leveraging knowledge distillation and self-training, which iteratively jointly trains two models in a teacher-student framework. The model leverages noised unlabeled data alongside labeled data to obtain better performance. We detail the setup we use as well as the various noising techniques in Appendix C. A vital aspect of our framework, however, is that we use unlabeled data from COVID-19. Concretely, in the inter-domain adaptation experiments, where we aim to transfer information from GoEmotions to \textsc{CovidEmo}, we use labeled data from GoEmotions alongside unlabeled data from \textsc{CovidEmo} (we make sure the model does not see any example from the test set). In the temporal setup, where we train on the first six months $\mathcal{F}_{tr}$ and test on the last six $\mathcal{F}_{ts}$, we use $\mathcal{F}_{tr}$ in conjunction with unlabeled data generated in the last six months.

\paragraph{Results.} We show the results obtained using Noisy Student training in Table \ref{tab:ssl}. \textbf{Our SSL technique bridges both the inter-domain and the in domain temporal performance gaps.} First, we note that our SSL-powered CT-BERT model trained on GoEmotions outperforms the plain CT-BERT by as much as $1\%$ in average macro F-1. Moreover, in our temporal transfer experiments, Noisy Student improves the performance of the model by $1.7\%$. These results are statistically significant, and emphasize that our method obtains better generalization performance and can be leveraged to produces better domain adaptation models.

\section{Related Work}

\paragraph{Emotion datasets.}
Emotion detection has been studied extensively with applications in music \cite{strapparava2012parallel}, social networks \cite{mohammad-2012-emotional,islam2019multi}, online news \cite{5360297}, and literature \cite{liu2019dens}. All these domains can be examined with the help of large curated datasets. These datasets are created using automated approaches such as distant supervision  \cite{wang2012harnessing,abdul-mageed-ungar-2017-emonet}, while others are manually labeled using crowdsourcing \cite{aman2007identifying,poria2018meld,liu2019dens,sosea2020canceremo,demszky2020goemotions,desai2020detecting}. In this work, we resort to the latter and create {\sc CovidEmo}, a dataset of $2,847$ tweets annotated with the Pluchik-8 emotions.

\paragraph{Emotions detection methods.}
In the early stages of emotion detection, most approaches used feature-based methods, which usually leveraged hand-crafted lexicons, such as EmoLex \cite{emolex} or the Valance Arousal Lexicon \cite{vad-acl2018}. These features were subsequently used to build classifiers such as Logistic Regression or SVMs. However, due to the recent advancements in deep learning as well as large pre-trained language models, all state-of-the-art approaches employ BERT-based \cite{devlin2019bert} classifiers.

\paragraph{COVID-19 emotion analysis.}
Since the emergence of the pandemic, numerous studies have been carried out on social media networks to understand COVID-19 and its effects on the larger population. \citet{ils-etal-2021-changes} annotated $2.3K$ German and English tweets for the expression of solidarity and used it to carry out an analysis into the expression of solidarity over time. On the other hand, \citet{saakyan-etal-2021-covid} annotated a dataset for detecting general misinformation in the pandemic. Sentiment analysis and emotion detection on social media during COVID-19 have seen tremendous popularity as well \cite{beck-etal-2021-investigating,KABIR2021100135,adikari_achini,choudrie_j,10.3389/fpsyg.2020.02252,calbi_m} due to the ability to provide vital information into the social aspects and the overall dynamics of the population. In this paper, however, we annotate {\sc CovidEmo}, a dataset of fine-grained emotions and employ a comprehensive analysis into cross-domain and temporal generalization of large pretrained language models.

\section{Conclusion}
We present {\sc CovidEmo}, a dataset of tweets annotated with perceived Plutchik-8 emotions. Using this dataset, we reveal emotion distributions and associations that are distinctive from prior studies on disaster-related emotion annotation and detection. We further show that models trained on other emotion datasets transfer poorly. Additionally, we indicate that models transfer poorly when trained on different temporal slices of an event such as COVID-19. Next, we conduct a comprehensive analysis of the temporal and inter-domain gaps to offer a better understanding of why models transfer poorly. As a potential solution to bridge these gaps and offer a more reliable disaster response, we leverage the large amount of readily available data alongside semi-supervised learning techniques.

\section*{Acknowledgements}
This work was partially supported by NSF Grants IIS-2107524, IIS-1850153, IIS-2107487, and BigData-1912887.
We acknowledge the Texas Advanced Computing Center (TACC) at The University of Texas at Austin for providing HPC resources that have contributed to the research results reported within this paper. The computation for this project was also performed on
Amazon Web Services through a research grant to UIC.

\bibliography{anthology,custom}

\begin{thebibliography}{48}
\expandafter\ifx\csname natexlab\endcsname\relax\def\natexlab#1{#1}\fi

\bibitem[{Abdul-Mageed and Ungar(2017)}]{abdul-mageed-ungar-2017-emonet}
Muhammad Abdul-Mageed and Lyle Ungar. 2017.
\newblock {EmoNet: Fine-Grained Emotion Detection with Gated Recurrent Neural
  Networks}.
\newblock In \emph{Proceedings of the 55th Annual Meeting of the Association
  for Computational Linguistics (Volume 1: Long Papers)}, pages 718--728.

\bibitem[{Adikari et~al.(2021)Adikari, Nawaratne, De~Silva, Ranasinghe,
  Alahakoon, Alahakoon et~al.}]{adikari_achini}
Achini Adikari, Rashmika Nawaratne, Daswin De~Silva, Sajani Ranasinghe, Oshadi
  Alahakoon, Damminda Alahakoon, et~al. 2021.
\newblock Emotions of covid-19: Content analysis of self-reported information
  using artificial intelligence.
\newblock \emph{Journal of Medical Internet Research}, 23(4):e27341.

\bibitem[{Aman and Szpakowicz(2007)}]{aman2007identifying}
Saima Aman and Stan Szpakowicz. 2007.
\newblock Identifying expressions of emotion in text.
\newblock In \emph{International Conference on Text, Speech and Dialogue},
  pages 196--205.

\bibitem[{{Bao} et~al.(2009){Bao}, {Xu}, {Zhang}, {Yan}, {Su}, {Han}, and
  {Yu}}]{5360297}
S.~{Bao}, S.~{Xu}, L.~{Zhang}, R.~{Yan}, Z.~{Su}, D.~{Han}, and Y.~{Yu}. 2009.
\newblock Joint emotion-topic modeling for social affective text mining.
\newblock In \emph{2009 Ninth IEEE International Conference on Data Mining},
  pages 699--704.

\bibitem[{Beck et~al.(2021)Beck, Lee, Viehmann, Maurer, Quiring, and
  Gurevych}]{beck-etal-2021-investigating}
Tilman Beck, Ji-Ung Lee, Christina Viehmann, Marcus Maurer, Oliver Quiring, and
  Iryna Gurevych. 2021.
\newblock Investigating label suggestions for opinion mining in {G}erman
  covid-19 social media.
\newblock In \emph{Proceedings of the 59th Annual Meeting of the Association
  for Computational Linguistics and the 11th International Joint Conference on
  Natural Language Processing (Volume 1: Long Papers)}, pages 1--13.

\bibitem[{Blei et~al.(2003)Blei, Ng, and Jordan}]{10.5555/944919.944937}
David~M Blei, Andrew~Y Ng, and Michael~I Jordan. 2003.
\newblock Latent dirichlet allocation.
\newblock \emph{The Journal of Machine Learning Research}, 3:993--1022.

\bibitem[{Cachola et~al.(2018)Cachola, Holgate, Preo{\c{t}}iuc-Pietro, and
  Li}]{cachola-etal-2018-expressively}
Isabel Cachola, Eric Holgate, Daniel Preo{\c{t}}iuc-Pietro, and Junyi~Jessy Li.
  2018.
\newblock Expressively vulgar: The socio-dynamics of vulgarity and its effects
  on sentiment analysis in social media.
\newblock In \emph{Proceedings of the 27th International Conference on
  Computational Linguistics}, pages 2927--2938.

\bibitem[{Calbi et~al.(2021)Calbi, Langiulli, Ferroni, Montalti, Kolesnikov,
  Gallese, and Umilt{\`a}}]{calbi_m}
Marta Calbi, Nunzio Langiulli, Francesca Ferroni, Martina Montalti, Anna
  Kolesnikov, Vittorio Gallese, and Maria~Alessandra Umilt{\`a}. 2021.
\newblock The consequences of covid-19 on social interactions: an online study
  on face covering.
\newblock \emph{Scientific Reports}, 11(1):1--10.

\bibitem[{Chen et~al.(2020)Chen, Lerman, and Ferrara}]{emilychendata}
Emily Chen, Kristina Lerman, and Emilio Ferrara. 2020.
\newblock Tracking social media discourse about the covid-19 pandemic:
  Development of a public coronavirus twitter data set.
\newblock \emph{JMIR Public Health Surveill}, 6(2):e19273.

\bibitem[{Choudrie et~al.(2021)Choudrie, Patil, Kotecha, Matta, and
  Pappas}]{choudrie_j}
Jyoti Choudrie, Shruti Patil, Ketan Kotecha, Nikhil Matta, and Ilias Pappas.
  2021.
\newblock Applying and understanding an advanced, novel deep learning approach:
  A covid 19, text based, emotions analysis study.
\newblock \emph{Information Systems Frontiers}, pages 1--35.

\bibitem[{Demszky et~al.(2020)Demszky, Movshovitz-Attias, Ko, Cowen, Nemade,
  and Ravi}]{demszky2020goemotions}
Dorottya Demszky, Dana Movshovitz-Attias, Jeongwoo Ko, Alan Cowen, Gaurav
  Nemade, and Sujith Ravi. 2020.
\newblock {G}o{E}motions: A dataset of fine-grained emotions.
\newblock In \emph{Proceedings of the 58th Annual Meeting of the Association
  for Computational Linguistics}, pages 4040--4054.

\bibitem[{Dennis et~al.(2006)Dennis, Kunkel, Woods, and
  Schrodt}]{dennis2006making}
Michael~Robert Dennis, Adrianne Kunkel, Gillian Woods, and Paul Schrodt. 2006.
\newblock {Making Sense of New Orleans Flood Trauma Recovery: Ethics, Research
  Design, and Policy Considerations for Future Disasters}.
\newblock \emph{Analyses of Social Issues and Public Policy}, 6(1):191--213.

\bibitem[{Desai et~al.(2020)Desai, Caragea, and Li}]{desai2020detecting}
Shrey Desai, Cornelia Caragea, and Junyi~Jessy Li. 2020.
\newblock Detecting perceived emotions in hurricane disasters.
\newblock In \emph{Proceedings of the 58th Annual Meeting of the Association
  for Computational Linguistics}, pages 5290--5305.

\bibitem[{Devlin et~al.(2019)Devlin, Chang, Lee, and
  Toutanova}]{devlin2019bert}
Jacob Devlin, Ming-Wei Chang, Kenton Lee, and Kristina Toutanova. 2019.
\newblock Bert: Pre-training of deep bidirectional transformers for language
  understanding.
\newblock In \emph{Proceedings of the 2019 Conference of the North American
  Chapter of the Association for Computational Linguistics: Human Language
  Technologies, Volume 1 (Long and Short Papers)}, pages 4171--4186.

\bibitem[{Fellbaum(2012)}]{fellbaum2012wordnet}
Christiane Fellbaum. 2012.
\newblock Wordnet.
\newblock \emph{The encyclopedia of applied linguistics}.

\bibitem[{Fox(2020)}]{q4-vaccine}
Maggie Fox. 2020.
\newblock \href
  {https://www.cnn.com/2020/11/24/health/covid-vaccines-design-explained/index.html}
  {Here's a look at how the different coronavirus vaccines work}.
\newblock \emph{CNN}.

\bibitem[{Fraustino et~al.(2012)Fraustino, Liu, and Jin}]{fraustino2012social}
Julia~Daisy Fraustino, Brooke~Fisher Liu, and Yan~Xian Jin. 2012.
\newblock {Social Media Use During Disasters: A Review of the Knowledge Base
  and Gaps}.
\newblock \emph{National Consortium for the Study of Terrorism and Responses to
  Terrorism}.

\bibitem[{Ils et~al.(2021)Ils, Liu, Grunow, and Eger}]{ils-etal-2021-changes}
Alexandra Ils, Dan Liu, Daniela Grunow, and Steffen Eger. 2021.
\newblock Changes in {E}uropean solidarity before and during {COVID}-19:
  Evidence from a large crowd- and expert-annotated {T}witter dataset.
\newblock In \emph{Proceedings of the 59th Annual Meeting of the Association
  for Computational Linguistics and the 11th International Joint Conference on
  Natural Language Processing (Volume 1: Long Papers)}, pages 1623--1637.

\bibitem[{Islam et~al.(2019)Islam, Mercer, and Xiao}]{islam2019multi}
Jumayel Islam, Robert~E Mercer, and Lu~Xiao. 2019.
\newblock Multi-channel convolutional neural network for twitter emotion and
  sentiment recognition.
\newblock In \emph{Proceedings of the 2019 Conference of the North American
  Chapter of the Association for Computational Linguistics: Human Language
  Technologies, Volume 1 (Long and Short Papers)}, pages 1355--1365.

\bibitem[{Joulin et~al.(2017)Joulin, Grave, Bojanowski, and Mikolov}]{fasttext}
Armand Joulin, {\'E}douard Grave, Piotr Bojanowski, and Tom{\'a}{\v{s}}
  Mikolov. 2017.
\newblock Bag of tricks for efficient text classification.
\newblock In \emph{Proceedings of the 15th Conference of the European Chapter
  of the Association for Computational Linguistics: Volume 2, Short Papers},
  pages 427--431.

\bibitem[{Kabir and Madria(2021)}]{KABIR2021100135}
Md.~Yasin Kabir and Sanjay Madria. 2021.
\newblock Emocov: Machine learning for emotion detection, analysis and
  visualization using covid-19 tweets.
\newblock \emph{Online Social Networks and Media}, 23:100135.

\bibitem[{Kaplan(2021)}]{q1-stimulus}
Thomas Kaplan. 2021.
\newblock \href
  {https://www.nytimes.com/2021/03/07/us/politics/whats-in-the-stimulus-bill.html}
  {What's in the stimulus bill? a guide to where the \$1.9 trillion is going}.
\newblock \emph{The New York Times}.

\bibitem[{Liu et~al.(2019)Liu, Osama, and De~Andrade}]{liu2019dens}
Chen Liu, Muhammad Osama, and Anderson De~Andrade. 2019.
\newblock {DENS}: A dataset for multi-class emotion analysis.
\newblock In \emph{Proceedings of the 2019 Conference on Empirical Methods in
  Natural Language Processing and the 9th International Joint Conference on
  Natural Language Processing (EMNLP-IJCNLP)}, pages 6293--6298.

\bibitem[{Mohammad(2012)}]{mohammad-2012-emotional}
Saif Mohammad. 2012.
\newblock {{\#}Emotional Tweets}.
\newblock In \emph{*{SEM} 2012: The First Joint Conference on Lexical and
  Computational Semantics {--} Volume 1: Proceedings of the main conference and
  the shared task, and Volume 2: Proceedings of the Sixth International
  Workshop on Semantic Evaluation ({S}em{E}val 2012)}, pages 246--255.

\bibitem[{Mohammad and Kiritchenko(2015)}]{mohammad2015using}
Saif Mohammad and Svetlana Kiritchenko. 2015.
\newblock {Using Hashtags to Capture Fine Emotion Categories from Tweets}.
\newblock \emph{Computational Intelligence}, 31(2):301--326.

\bibitem[{Mohammad(2018)}]{vad-acl2018}
Saif~M. Mohammad. 2018.
\newblock Obtaining reliable human ratings of valence, arousal, and dominance
  for 20,000 english words.
\newblock In \emph{Proceedings of The Annual Conference of the Association for
  Computational Linguistics (ACL)}, Melbourne, Australia.

\bibitem[{Mohammad and Turney(2013)}]{emolex}
Saif~M Mohammad and Peter~D Turney. 2013.
\newblock Crowdsourcing a word--emotion association lexicon.
\newblock \emph{Computational intelligence}, 29(3):436--465.

\bibitem[{M{\"u}ller et~al.(2020)M{\"u}ller, Salath{\'e}, and
  Kummervold}]{muller2020covid}
Martin M{\"u}ller, Marcel Salath{\'e}, and Per~E Kummervold. 2020.
\newblock Covid-twitter-bert: A natural language processing model to analyse
  covid-19 content on twitter.
\newblock \emph{arXiv preprint arXiv:2005.07503}.

\bibitem[{Nguyen et~al.(2020)Nguyen, Vu, and Nguyen}]{bertweet}
Dat~Quoc Nguyen, Thanh Vu, and Anh~Tuan Nguyen. 2020.
\newblock {BERTweet: A pre-trained language model for English Tweets}.
\newblock In \emph{Proceedings of the 2020 Conference on Empirical Methods in
  Natural Language Processing: System Demonstrations}, pages 9--14.

\bibitem[{Nilsen and Zhou(2020)}]{q2-stimulus}
Ella Nilsen and Li~Zhou. 2020.
\newblock \href
  {https://www.vox.com/2020/5/12/21254397/next-coronavirus-stimulus-package-democrats-heroes-ac}
  {Democrats\' \$3 trillion opening bid for the next stimulus package,
  explained}.
\newblock \emph{Vox}.

\bibitem[{Nye and Nenkova(2015)}]{nye2015identification}
Benjamin Nye and Ani Nenkova. 2015.
\newblock Identification and characterization of newsworthy verbs in world
  news.
\newblock In \emph{Proceedings of the 2015 Conference of the North American
  Chapter of the Association for Computational Linguistics: Human Language
  Technologies}, pages 1440--1445.

\bibitem[{Oliphant(2020)}]{q4-election}
James Oliphant. 2020.
\newblock \href
  {https://www.reuters.com/article/global-poy-usa-election-idUSKBN28K1FU} {U.s.
  election year shaped by pandemic and trump's defiance}.
\newblock \emph{Reuters}.

\bibitem[{Organization(2020)}]{q1-who}
World~Health Organization. 2020.
\newblock Novel coronavirus (2019-ncov): situation report, 19.
\newblock Technical documents.

\bibitem[{Palen and Anderson(2016)}]{palen2016crisis}
Leysia Palen and Kenneth~M Anderson. 2016.
\newblock Crisis informatics-new data for extraordinary times.
\newblock \emph{Science}, 353(6296):224--225.

\bibitem[{Plutchik(2001)}]{plutchik2001nature}
Robert Plutchik. 2001.
\newblock The nature of emotions: Human emotions have deep evolutionary roots,
  a fact that may explain their complexity and provide tools for clinical
  practice.
\newblock \emph{American scientist}, 89(4):344--350.

\bibitem[{Poria et~al.(2019)Poria, Hazarika, Majumder, Naik, Cambria, and
  Mihalcea}]{poria2018meld}
Soujanya Poria, Devamanyu Hazarika, Navonil Majumder, Gautam Naik, Erik
  Cambria, and Rada Mihalcea. 2019.
\newblock {MELD}: A multimodal multi-party dataset for emotion recognition in
  conversations.
\newblock In \emph{Proceedings of the 57th Annual Meeting of the Association
  for Computational Linguistics}, pages 527--536.

\bibitem[{Saakyan et~al.(2021)Saakyan, Chakrabarty, and
  Muresan}]{saakyan-etal-2021-covid}
Arkadiy Saakyan, Tuhin Chakrabarty, and Smaranda Muresan. 2021.
\newblock {COVID}-fact: Fact extraction and verification of real-world claims
  on {COVID}-19 pandemic.
\newblock In \emph{Proceedings of the 59th Annual Meeting of the Association
  for Computational Linguistics and the 11th International Joint Conference on
  Natural Language Processing (Volume 1: Long Papers)}, pages 2116--2129.

\bibitem[{Scarpina(2020)}]{10.3389/fpsyg.2020.02252}
Federica Scarpina. 2020.
\newblock Detection and recognition of fearful facial expressions during the
  coronavirus disease (covid-19) pandemic in an italian sample: An online
  experiment.
\newblock \emph{Frontiers in Psychology}, 11:2252.

\bibitem[{Schulz et~al.(2013)Schulz, Thanh, Paulheim, and
  Schweizer}]{Schulz2013AFS}
Axel Schulz, Tung~Dang Thanh, Heiko Paulheim, and Immanuel Schweizer. 2013.
\newblock {A Fine-Grained Sentiment Analysis Approach for Detecting Crisis
  Related Microposts}.
\newblock In \emph{Information Systems for Crisis Response and Management
  (ISCRAM)}.

\bibitem[{Slater(2021)}]{q2-india}
Joanna Slater. 2021.
\newblock \href
  {https://www.washingtonpost.com/world/2021/05/01/india-coronavirus/} {India
  sets pandemic record with more than 400,000 new cases; fauci says crisis is
  `like a war'}.
\newblock \emph{The Washington Post}.

\bibitem[{Sosea and Caragea(2020)}]{sosea2020canceremo}
Tiberiu Sosea and Cornelia Caragea. 2020.
\newblock Canceremo: A dataset for fine-grained emotion detection.
\newblock In \emph{Proceedings of the 2020 Conference on Empirical Methods in
  Natural Language Processing (EMNLP)}, pages 8892--8904.

\bibitem[{Sprunt and Turner(2020)}]{q3-school}
Barbara Sprunt and Cory Turner. 2020.
\newblock \href
  {https://www.npr.org/2020/07/08/888898194/trump-blasts-expensive-cdc-guidelines-for-reopening-schools}
  {White house stumbles over how best to reopen schools, as trump blasts cdc
  guidance}.
\newblock \emph{NPR}.

\bibitem[{Strapparava et~al.(2012)Strapparava, Mihalcea, and
  Battocchi}]{strapparava2012parallel}
Carlo Strapparava, Rada Mihalcea, and Alberto Battocchi. 2012.
\newblock A parallel corpus of music and lyrics annotated with emotions.
\newblock In \emph{Proceedings of the Eighth International Conference on
  Language Resources and Evaluation (LREC'12)}, pages 2343--2346.

\bibitem[{Volkova and Bachrach(2016)}]{volkova-bachrach-2016-inferring}
Svitlana Volkova and Yoram Bachrach. 2016.
\newblock {Inferring Perceived Demographics from User Emotional Tone and
  User-Environment Emotional Contrast}.
\newblock In \emph{Proceedings of the 54th Annual Meeting of the Association
  for Computational Linguistics (Volume 1: Long Papers)}, pages 1567--1578.

\bibitem[{Wang et~al.(2012)Wang, Chen, Thirunarayan, and
  Sheth}]{wang2012harnessing}
Wenbo Wang, Lu~Chen, Krishnaprasad Thirunarayan, and Amit~P Sheth. 2012.
\newblock Harnessing twitter ``big data" for automatic emotion identification.
\newblock In \emph{2012 International Conference on Privacy, Security, Risk and
  Trust and 2012 International Confernece on Social Computing}, pages 587--592.

\bibitem[{Wee et~al.(2020)Wee, Mcneil, and Hernández}]{q2-pandemic}
Sui-lee Wee, Donald~G. Mcneil, and Javier~C. Hernández. 2020.
\newblock \href
  {https://www.nytimes.com/2020/01/30/health/coronavirus-world-health-organization.html}
  {W.h.o. declares global emergency as wuhan coronavirus spreads}.
\newblock \emph{The New York Times}.

\bibitem[{Xie et~al.(2020)Xie, Luong, Hovy, and Le}]{xie2020self}
Qizhe Xie, Minh-Thang Luong, Eduard Hovy, and Quoc~V Le. 2020.
\newblock Self-training with noisy student improves imagenet classification.
\newblock In \emph{Proceedings of the IEEE/CVF Conference on Computer Vision
  and Pattern Recognition}, pages 10687--10698.

\bibitem[{Zhu et~al.(2015)Zhu, Kiros, Zemel, Salakhutdinov, Urtasun, Torralba,
  and Fidler}]{zhu2015aligning}
Yukun Zhu, Ryan Kiros, Rich Zemel, Ruslan Salakhutdinov, Raquel Urtasun,
  Antonio Torralba, and Sanja Fidler. 2015.
\newblock Aligning books and movies: Towards story-like visual explanations by
  watching movies and reading books.
\newblock In \emph{Proceedings of the IEEE international conference on computer
  vision}, pages 19--27.

\end{thebibliography}
\bibliographystyle{acl_natbib}

\appendix

\section{Hyperparameters Used}
In all our experiments, we found that a batch size of $16$ works best. Additionally, we indicate in Table \ref{tab:learning_rates} the best learning rates for our models. We refrain from showing the best learning rates on HurricaneEmo due to low performance, high variance of the results.

\begin{table*}[t]
\setlength{\tabcolsep}{7pt}
\small
\centering 
\begin{tabular}{rcccccccc}
\toprule
& ang & ant & dis & fea & joy & sad & sur & tru\\
\midrule
\textsc{bert-goemotions} & $3e-05$ & $5e-05$ & $7e-0$5 & $5e-05$ & $5e-05$ & $5e-05$ & $3e-05$ & $5e-05$\\
\textsc{tweetbert-goemotions} & $3e-05$ & $3e-05$ & $5e-05$ & $5e-05$ & $5e-05$ & $7e-05$ & $7e-05$ & $5e-05$\\
\textsc{ctbert-goemotions} & $5e-0$5 & $7e-05$ & $3e-05$ & $5e-05$ & $5e-05$ & $3e-05$ & $3e-05$ & $5e-05$\\
\textsc{ctbert-goemotions-ssl} & $3e-05$ & $1e-05$ & $5e-05$ & $3e-05$ & $7e-05$ & $7e-05$ & $5e-05$ & $5e-05$\\
\textsc{ctbert-}$\mathcal{F}_{tr}$ & $5e-05$ & $7e-05$ & $5e-05$ & $5e-05$ & $3e-05$ & $3e-05$ & $5e-05$ & $5e-05$\\
\textsc{ctbert-}$\mathcal{F}_{tr}$\textsc{-ssl} & $5e-05$ & $5e-05$ & $5e-05$ & $7e-05$ & $7e-05$ & $5e-05$ & $5e-05$ & $5e-05$\\
\bottomrule
\end{tabular}
\caption{Best Learning Rates for for our models.}
\label{tab:learning_rates}
\end{table*}

\section{Hyperparameter Search Space}

For each emotion, we investigate with batch sizes in the set [$8$, $16$, $32$], and train for up to $5$ epochs with early stopping. In terms of learning rates, we follow the best practices from the orignal BERT paper and explore learning rates around $5e-5$. Specifically, we experiment with values in the range $1e-5$ -> $9e-5$ with steps of $2e-5$. Although hyperparameter tuning is quite expensive computationally ($15$ runs per emotion per model), we found that the default BERT setup (5e-5 learning rate and a batch size of $32$) works within $0.5\%$ F-1 of the best model.

\section{Semi-supervised Learning}
\vspace{6mm}
 Noisy Student training \cite{xie2020self} leverages knowledge distillation (KD) and self-training to iteratively train two models in a teacher-student framework. The framework trains the student in traditional KD fashion, matching its predictions to those of the teacher. Concretely the training loss is:

\begin{align*}
\centering
\mathcal{L} = \sum_{(x^{i}, y^{i}) \in \mathcal{C}_{tr}} l(f_{\tau}(x^{i}), f_{\tau^{'}}(x^{i})),
\end{align*}

\noindent where $\mathcal{D}$ is the training dataset, $l$ is the cross-entropy loss, and $f_{\tau}$ and $f_{\tau^{'}}$ are the student and the teacher models, respectively. We note one vital particularity of this framework: The student is trained using noised input examples. In the orginal paper, the authors also use a larger network for the student, but we noticed here that using equal-sized architectures works well enough. Leveraging noised inputs, Noisy Student exposes the student to more difficult learning environments, and usually leads to an increased performance compared to the teacher. To add noise to our input examples, we use two approaches: a) {\em Synonym replacement:} We replace between one and three words in a tweet with its synonym using the WordNet English lexical database \cite{fellbaum2012wordnet}; b) {\em Back-translation:} We use {back-translation}, and experiment with different levels of noise corresponding to different translation chain lengths (e.g., English-French-Spanish-English). Smaller chain lengths lead to less noise, while increasing the length of the chain produces examples with significantly more noise. For each unlabeled example, we sample uniformly a chain length in the range $1$->$10$, and use the following languages for translation: Russian, French, Spanish, Italian, and German.


\end{document}